\documentclass[runningheads]{llncs}
\usepackage[T1]{fontenc}
\usepackage{graphicx}
\usepackage{amsmath}
\usepackage{amssymb}
\usepackage{algorithm}
\usepackage{algpseudocode}
\usepackage{bbm}
\usepackage{physics}
\usepackage{subcaption}
\usepackage{xcolor}
\usepackage{hyperref}
\newtheorem{defn}{Definition}

\begin{document}
\title{Enhance Exploration in Safe Reinforcement Learning with Contrastive Representation Learning}
\titlerunning{Enhance Exploration in SafeRL with Contrastive Representation Learning}
%
%
\author{Duc Kien Doan \and
Bang Giang Le \and
Viet Cuong Ta$^*$}
\authorrunning{Doan et al.}
%
\institute{HMI Laboratory \\ 
VNU University of Engineering and Technology, Hanoi, Vietnam 
\\ $^*$Corresponding Author: \email{cuongtv@vnu.edu.vn}
}
\maketitle              
\begin{abstract}

In safe reinforcement learning, agent needs to balance between exploration actions and safety constraints.
Following this paradigm, domain transfer approaches learn a prior Q-function from the related environments to prevent unsafe actions.
However, because of the large number of false positives, some safe actions are never executed, leading to inadequate exploration in sparse-reward environments.
In this work, we aim to learn an efficient state representation to balance the exploration and safety-prefer action in a sparse-reward environment.
Firstly, the image input is mapped to latent representation by an auto-encoder.
A further contrastive learning objective is employed to distinguish safe and unsafe states.
In the learning phase, the latent distance is used to construct an additional safety check, which allows the agent to bias the exploration if it visits an unsafe state.
To verify the effectiveness of our method, the experiment is carried out in three navigation-based MiniGrid environments.
The result highlights that our method can explore the environment better while maintaining a good balance between safety and efficiency.

\end{abstract}
\section{Introduction}
Reinforcement learning (RL) has been extensively used to solve problems in various domains such as robotics and video games by learning a policy that maximizes cumulative rewards \cite{dqn,silver2017mastering}.
However, in many applications, the policy is also required to avoid hazards, such as in driving tasks \cite{scheel2022urban} {and energy management \cite{NAKABI2021100413}}. Such requirement is of particular relevance to safety-critical tasks in the real world, where the cost of violation is prohibitive \cite{kamran2021minimizing}.
In other words, a RL agent has to solve two problems which are reaching the goal and taking preventive actions simultaneously.

{One of the main challenges in safe RL is balancing between rewards and adherence to safety constraints. Agents without safety awareness are usually able to attain high rewards by taking unsafe actions, in contrast to safe RL agents, which are able to avoid safe violations at the cost of lower rewards \cite{Achiam2019BenchmarkingSE}.
In constrained-based approaches, the standard Markov Decision Process (MDP) is alternated with a safety cost, which then could by solved by Lagrange multiplier \cite{liang2018acceleratedprimaldualpolicyoptimization} or other methods \cite{achiam2017constrainedpolicyoptimization}.
Shielding-based approaches \cite{alshiekh2017safereinforcementlearningshielding} attempt to predict the unsafe actions, and use this information to prevent the exploration process violate the safety conditions.}
To extract necessary safety knowledge from environments, Karimpanal et. al \cite{karimpanal2020} propose the usage of domain priors, where related environments can be used to train a safety critic.
{Other methods can also be employed to train the safety critic, such as by constraining the actions of a policy \cite{srinivasan2020learningsafedeeprl}.}
However, in a sparse-reward environment, the trained critic is too conservative, which hinders exploration and completing the task.
From an exploitation-exploration perspective in standard RL settings, strategies like $\epsilon$-greedy exploration \cite{ding2023incremental} or entropy-based exploration \cite{han2021diversity} can be used to control the policy's behaviors.
{In addition, }latent-based methods search for an efficient state representation which could reflect the underlying MDP structures and use this representation to enhance the exploration process \cite{raileanu2020ride,le2023structural}.


Motivated by latent-based exploration strategies, we propose a two-stage shielding method that aims to balance exploration and adhering to safety constraints.
Using the domain prior knowledge, we train an auto-encoder with contrastive learning to map raw observations to a latent space, together with a safety critic based on transferable domain priors \cite{karimpanal2020}.
By learning this representation, the agent gains an intrinsic understanding of the safety landscape of its environment, allowing it to better navigate complex, high-dimensional observation spaces. In the second stage, once safe and unsafe states are identified, the agent assesses action safety through a critic, which identifies and replaces unsafe actions with safer alternatives. By decoupling state and action safety, our approach minimizes exploration bias, concentrating it only within unsafe regions and enabling the agent to explore safely and efficiently.
We test our approach in three goal-based navigation environments with safety setting.
Our method with latent learning outperforms other baselines in training efficiency. {Further analysis reveals that our approach achieves a better balance between exploration and safety compared to other methods.}

\section{Related Work}
\textbf{Safe reinforcement learning.} Several methods have been proposed to solve the problem of safe reinforcement learning. There are two main approaches, namely modifying the optimization criterion and modifying the exploration process \cite{JMLR:v16:garcia15a}. The former typically formulates the problem in a Constrained Markov Decision Process (CMDP) \cite{altman2021constrained} and incorporates safety constraints in policy optimization to ensure the cumulative episodic costs are lower than a threshold. Some of them optimize the policy with a cost constraint using Lagrange multiplier \cite{liang2018acceleratedprimaldualpolicyoptimization}, while others directly optimize the policy to ensure safety \cite{achiam2017constrainedpolicyoptimization}.

{Shielding-based approach} \cite{alshiekh2017safereinforcementlearningshielding}, typically intervene in the exploration process by modifying actions that lead to unfavorable conditions.
SAILR \cite{wagener2021safereinforcementlearningusing} defines intervention rules that are based on cost advantage functions. When this function is higher than some threshold, the agent executes a backup policy that is assumed to be safe from the initial state with high probability. Similarly, \cite{srinivasan2020learningsafedeeprl} pre-trains a safety critic to evaluate unsafe conditions and choose safe actions.
{Karimpanal et al. \cite{karimpanal2020} train a safety critic to recognize unsafe actions based on pre-trained Q-functions in environments of the same domain.}
 
\textbf{Representation learning.} In standard RL settings, representation learning can be used to learn a lower-dimensional embedding of state or action spaces. This makes the learning more sample efficient and speeds up training, especially in environments with image observations. 
Agent's exploration behaviors can also be enhanced by latent-based intrinsic rewards \cite{raileanu2020ride,le2023structural}.
To learn a good representation, various auxiliary tasks have been employed such as reconstruction loss \cite{yarats2020improvingsampleefficiencymodelfree} or contrastive learning \cite{liu2021returnbasedcontrastiverepresentationlearning}. SLAC \cite{lee2020stochasticlatentactorcriticdeep} and Hafner et al \cite{hafner2019learninglatentdynamicsplanning} learns a latent variable model that closely follows the transition dynamics. As for safe RL, SafeSLAC \cite{hogewind2023safe} extends SLAC to include a latent variable model for cost prediction, a safety critic and introduces a safety contraint in the policy objective.
{Cen et al. \cite{cen2024feasibilityconsistentrepresentationlearning} propose FCSRL, which} learns a representation of state by ensuring transition dynamics consistency and feasibility consistency. 

\section{Background}
\subsection{Problem formulation}
We formulate the problem as safe RL in a Partially Observable Markov Decision Process (POMDP). Formally, a POMDP is identified by a tuple $(\mathcal{S, O, A, P}, r, \gamma)$, where $\mathcal{S}$ stands for the state space, $\mathcal{A}$ stands for the action space.

Consider an agent interacting with the environment by sampling actions from policy $\pi$. In the partial observability setting, the agent can only use observation $o_t$ from the observation space $\mathcal{O}$ as the input for the policy. 
After taking an action $a_t$ sampled from the policy, the transition dynamic $\mathcal{P}(s_{t+1}|s_t, a_t)$ is used to sample the next state $s_{t+1}$. After this transition, it receives a reward $r_t$ generated by the function $r: \mathcal{S} \times \mathcal{A}\times\mathcal{S}\rightarrow\mathbb{R}$.


For the safety constraint definitions, we give the following definition of state safety and action safety based on previous work on safe RL \cite{hansSafeExploration} \cite{thomas2022safereinforcementlearningimagining}. Similar to \cite{thomas2022safereinforcementlearningimagining}, we assume that the set of states that are considered safety violations, $\mathcal{S}_{\mathrm{unsafe}}$ is specified.
\begin{defn}
        \label{def:safety}
	A state $s$ is said to be
	\begin{itemize}
		\item a \textbf{safety violation} if $s \in \mathcal{S}_{\mathrm{unsafe}}$.
		\item \textbf{undesirable} if $s \notin \mathcal{S}_{\mathrm{unsafe}}$ but there exists an action $a$ such that $\mathcal{P}(s'|s, a) > 0$ for some $s' \in \mathcal{S}_{\mathrm{unsafe}}$. 
		\item \textbf{safe} if it is neither a safety violation nor undesirable, and \textbf{unsafe} otherwise.
	\end{itemize}
\end{defn}
{In the case of \textbf{undesirable} states, the action that leads to safety violation is said to be \textbf{unsafe}}. By employing the undesirable property of states, we focus on training an observation encoder which can distinguish between undesirable states and safe ones.
Therefore, the agent can select actions which are bias to exploration if it is not visiting unsafe states.

\subsection{Safe exploration with a safety critic}
Using a safety critic is a common technique in many safe exploration methods, especially those modifying the exploration process. Typically, a critic $Q_{\mathrm{safe}}(s, a)$ is used to evaluate the safety of the action $a$ when executed in state $s$. If $Q_\mathrm{safe} < \epsilon$ where $\epsilon$ is a threshold, the action $a$ is declared unsafe and replaced by some other safe action. What characterizes each method is the way the safety critic is learned.
{In this work, we employ the proposed method in \cite{karimpanal2020} to learn a safety critic.}

\begin{figure}
    \centering
    \includegraphics[width=\linewidth]{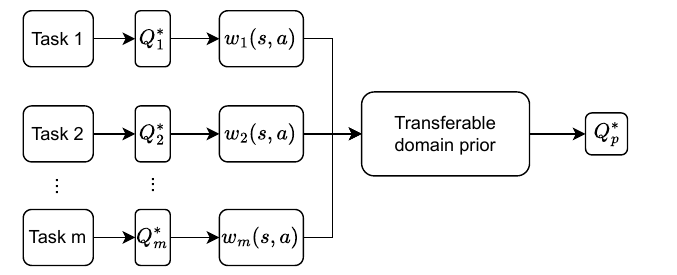}
    \caption{Illustration of training a prior Q-function to identify unsafe actions. {Given learned Q-functions of $m$ tasks in the same domain, weights are computed and assigned to each state-action pair. These pairs are then selected and used to construct a prior Q-function $Q^*_p$.}}
    \label{fig:transferable_domain_priors}
\end{figure}

More specifically, Karimpanal et. al \cite{karimpanal2020} rely on domain priors for safe exploration.
Firstly, a prior Q-function $Q_p$ is constructed to prevent the agent from taking undesirable actions.
Using the domain knowledge, the $Q_p$ is trained by aggregating $n$ optimal Q functions $Q^*_1, Q^*_2, \ldots, Q^*_n$ corresponding to $n$ tasks in the same domain. Particularly, based on these optimal Q functions, we can select state-action pairs that are consistently undesirable. To identify these pairs, the authors defined the scaled undesirability of a state-action pair $(s, a)$ with respect to task $i$ as:
\begin{equation}
	w_i(s, a) = \left| \frac{Q^*_i(s, a) - \max_{a'} Q^*_i(s, a')}{\max_{a'} Q_i^*(s,a)} \right|
\end{equation}
Since there are $n$ optimal Q functions, we can form an undesirablity vector $$W(s, a) = \{w_i(s, a)\}_{i=1}^n$$ and a derived probability distribution:
$$W'(s, a) = \left\{\frac{e^{w_i(s, a)}}{\sum_j{e^{w_j(s, a)}}}\right\}_{i=1}^n$$
To measure undesirablity consistency across tasks, we compute the product of mean of $W(s, a)$ and normalized entropy of $W'(s, a)$. Then, the state-action pairs with this value higher than a threshold are chosen, which means the corresponding transition reward are specified to be:
\begin{equation}
	r_p(s_c, a_c, s'_c) = \sum_{i=1}^N w_i'(s_c, a_c)[Q_i^*(s_c, a_c) - \gamma \max_{a' \in \mathcal{A}}Q_i^*(s'_c, a')] 
\end{equation}
This reward function is then used to train $Q_p$ (Figure \ref{fig:transferable_domain_priors}). After training, we can define unsafe actions as the ones that have $Q_p$ value smaller than the mean $Q_p$ for the corresponding state. As a result, during exploration, with some probability $\rho$, the agent checks if an action is unsafe. If this is the case, the agent chooses an action with $Q_p$ higher than average. {More formally, an action $a$ is considered unsafe in a state $s$ if in that state, ${\ensuremath{{Q_{P}^{*}}(s,a)<\underset{a'\in\mathcal{A}}{mean}{Q_{P}^{*}}(s,a')}}$.}


Although this method has shown positive impacts in reducing the number of safety violations, it has some major drawbacks, particularly in its definition of unsafe states. We consider \textit{unsafe actions} as the ones that lead to safety violations that the agent has to avoid. According to {the condition for marking an action as unsafe}, unless all $Q_p$ values are equal, which rarely happens, there always exists an unsafe action. However, in reality, this is not always the case. In safe states according to definition \ref{def:safety}, there should not be unsafe actions. By determining unsafe actions in safe states, the method inadvertently prevents some actions from being executed with high probability. This can impede the agent's exploration, which slows down learning, especially in sparse-reward environments. In these environments, only executing some actions can render the agent stuck in a safe region, making it unable to complete the task. In the following section, we propose a new method that tackles this problem. Our method adds a safety check for states, which prevents exploration in safe states from being biased.

\section{Method}
The main idea of our method is to identify and replace unsafe actions with two safety critics: state safety check and action safety check. At each step, the agent first performs the former to check if the current state is safe. If this is the case, it proceeds with an action proposed by the policy. Otherwise, there could be an action that leads to a safety violation, which means the agent has to check if the action is safe. {If the action found to be unsafe, it is replaced by a safer alternative}. By separating state safety from action safety, we can ensure that exploration is only biased in undesirable states. As a result, the agent can maintain safety while still being able to explore freely in safe states, which is crucial in reaching the goal.

Our method consists of two phases: pre-training the safety critics and using the critics to bias exploration towards safety while training the main policy. The action safety critic is trained based on the transferable domain priors method.
Regarding the state safety critic, to distinguish between safe and unsafe states while only having access to observations, the idea is to map them to a manageable latent space that takes into account safety. We use contrastive learning coupled with an auto-encoder (AE) to map the observations to a latent space that separates safe and unsafe ones. {Then, during execution, the agent maintains a buffer that stores latent representation of unsafe observations. To determine the safety of a state, the agent compares the embedding of the corresponding observation and a batch sampled from the buffer.}


\subsection{Learning contrastive latent representation} \label{learn_ae}
\begin{figure}
    \centering
    \includegraphics[width=\linewidth]{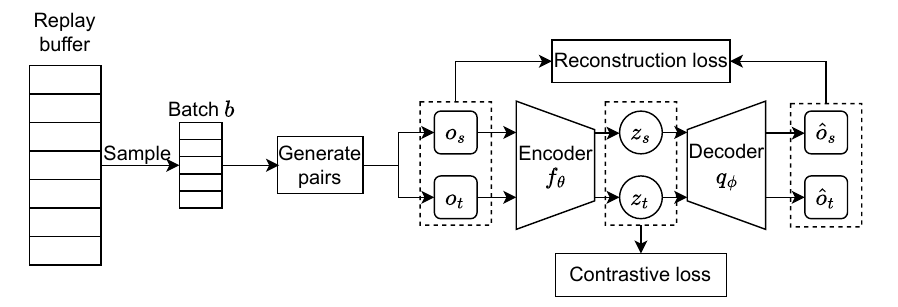}
    \caption{Learning a latent representation from past observations with contrastive learning.}
    \label{fig:method_phase1}
\end{figure}

To enable the agent to differentiate between unsafe and safe states {according to definition \ref{def:safety}}, we train an auto-encoder with encoder $f_\theta$ and decoder $g_\phi$.
{The training pipeline is illustrated in Figure \ref{fig:method_phase1}.}
The encoder $f_\theta$ takes as input the corresponding observation $o_t$ and outputs a latent representation $z_t$, where the Euclidean distance between two latent vectors are small if the corresponding states have the same safety characteristic and large otherwise. 
To enforce this constraint, we employ a contrastive loss function from deep metric learning \cite{contrastiveloss} which could reuse the safety information form the domain-prior knowledge. More formally, we define the function that denotes the safety of {an observation $o_s$ corresponding to a state $s$}:
\begin{equation}
    \mathrm{SAFE}({o_s}) = \begin{cases}
        1 & \text{if } s \text{ is safe}\\
        0 & \text{otherwise}
    \end{cases}
    \label{eq:AEobj}
\end{equation}
Then, the contrastive loss is defined as follows:
\begin{equation}
    \mathcal{L}_{\mathrm{c}}(o_s, o_t) =  \begin{cases}
        \norm{f_\theta(o_s) - f_\theta(o_t)}_2^2 & \text{if } \mathrm{SAFE}(o_s) = \mathrm{SAFE}(o_t)\\
        \max \left(0, \, \alpha - \norm{f_\theta(o_s) - f_\theta(o_t)}_2^2\right) & \text{otherwise}
    \end{cases}
\end{equation}

where the margin $\alpha$ is a hyperparameter that enforces the minimum value of the distance between dissimilar latent vector pairs. This loss function aims to minimize the distance between two latent embeddings of the same class and ensures that those in different classes are distant by setting a large value for $\alpha$. In addition to contrastive loss, to make the representation consistent, a reconstruction loss term is also employed, specifically the mean squared error (MSE) loss:
\begin{equation}
   \mathcal{L}_{\text{r}}(o_s) =  \norm{o_s - \hat{o}_s}^2_2
\end{equation}
where $\hat{o}_s = g_{\phi}(f_\theta(o_s))$.

To train the auto-encoder, we sample a batch $b$ from a replay buffer $\mathcal{B}$. This replay buffer is populated by storing observation $o_t$ along with the corresponding safety value $\mathrm{SAFE}(s_t)$ obtained in the next step. The training objective for the auto-encoder is defined as follows:
\begin{equation}
    \mathcal{L}_{AE} = {\omega}_1 \sum_{o_t \in {b}}\mathcal{L}_{\mathrm{r}}(o_t) + \omega_2 \sum_{o_s, o_t \in {b}} \mathcal{L}_{\mathrm{c}}(o_s ,o_t)
    \label{eq:AEobj}
\end{equation}

\subsection{Bias exploration with latent distance check}
\begin{figure}
	\centering
	\includegraphics[width=\linewidth]{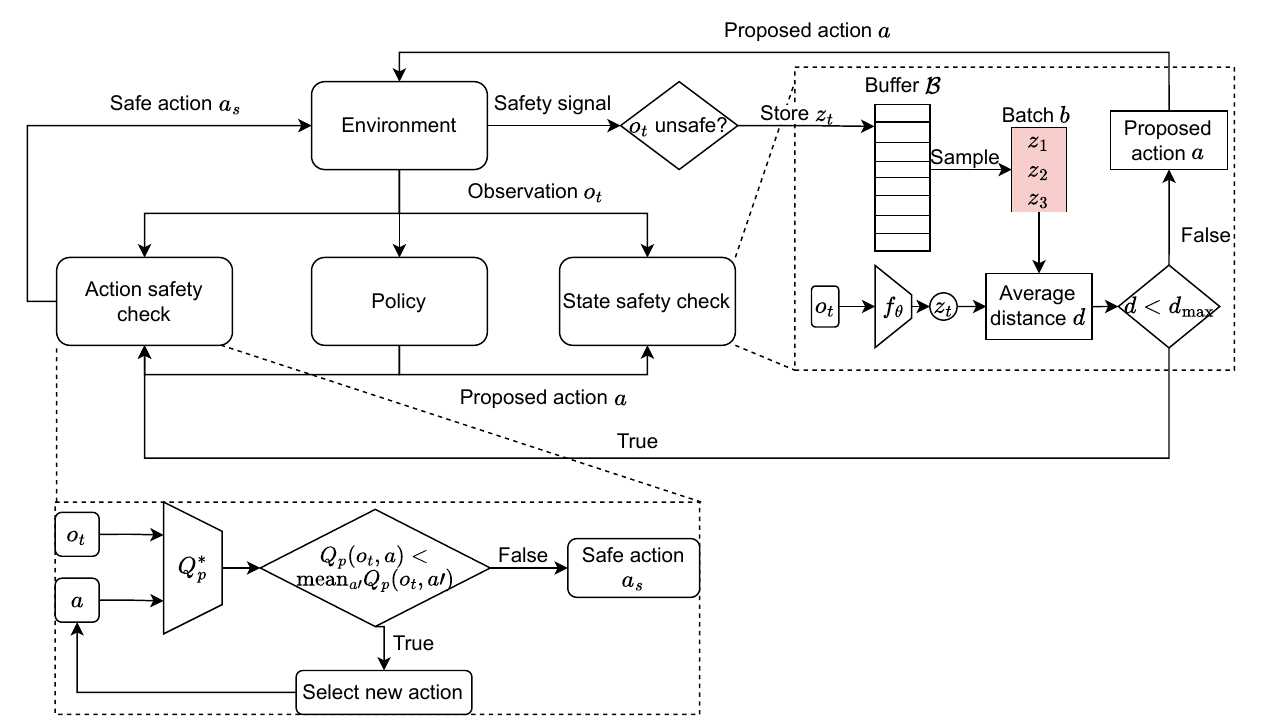}
        \caption{{Illustration of our method. Given an observation $o_t$ and proposed action $a$, the agent performs a state safety check based on the average distance from the current observation embedding and embeddings in the unsafe buffer $\mathcal{B}$. If the current state is safe, the proposed action is executed. Otherwise, the agent executes a safe action computed by the action safety check module.}}
	\label{fig:method_overview}
\end{figure}

After training an auto-encoder to map observations to the desired latent space, we can employ it to bias exploration towards safety. Suppose the exploration algorithm already proposes an action, the idea is to add a safety module that prevents the agent from taking unsafe actions. This module consists of two components: state safety check and action safety check, where the latter follows the transferable domain priors method in \cite{karimpanal2020}. As for the state safety check module, illustrated in Figure \ref{fig:method_overview}, at every step, the agent samples a batch $b$ from a buffer $\mathcal{B}$ that only stores the embeddings of past unsafe observations. Then, the agent compares the embedding of the current observation to those in $b$. If the difference is small, the current state is considered unsafe, which means the action safety check component is executed. At any step, if the agent visits an undesirable state, indicating an unsafe previous state, the embedding of the previous state is appended to the buffer. With the overview of the state safety component in mind, we now provide the details of each part.


\textbf{Unsafe embedding buffer.} We maintain a buffer $\mathcal{B}$ in an online fashion that stores embedding of unsafe states that the agent visits. For each transition tuple $(o_t, a_t, r_t, o_{t+1})$ at step $t$, we append the embedding $z_{t} = f_\theta(o_t)$ to $\mathcal{B}$ if the transition leads to unfavorable conditions. This can be death, low rewards, or high costs, etc. This buffer serves as a memory of past experiences that the agent can query from. The intuition is that the agent can ``remember'' unsafe states so that it can avoid them in the future.

\textbf{State safety condition.} Based on the property that embeddings of unsafe state are close in the latent space, we derive a rule to determine whether a particular state is safe. Given an observation $o_s$ and its embedding $z_s = f_\theta(o_s)$, we compare it to a batch $b$ sampled from the buffer $\mathcal{B}$. To be more specific, we compute the mean L2 distance between $z_s$ and every embedding in $b$:
\begin{equation}
    d = \frac{1}{|b|}\sum_{z_t \in b} \norm{z_s - z_t}_2
\end{equation}
If this distance is smaller than a threshold $d_{\max}$, which means $s$ is close to unsafe states in the latent space, $s$ is considered unsafe. In this case, the agent continues to check if the proposed action is unsafe according to definition \ref{def:safety}. 

{Algorithm \ref{alg:algorithm3} outlines the process of biasing exploration against undesirable behavior with the addition of state safety check. The distance $d$ is first computed between embeddings of the current state and that of unsafe states sampled from the buffer. Using this distance, the agent can check for state safety conditions. Then the agent proceeds to choose safe actions based on the prior Q-function.}

\begin{algorithm}[h]
\caption{{Apply state safety check to bias against undesirable exploration}}
\begin{algorithmic}[1]
\State \textbf{Input: }
\State Proposed exploratory action $a_{0}$, state $s$, optimal
$Q$- function of prior $Q_{P}^{*}$, probability of using priors
$\rho$, encoder $f_\theta(s)$, unsafe buffer $\mathcal{B}$, latent distance threshold $d_{\max}$.
\State \textbf{Output: }selected action $a$\textbf{ }
\State Sample a batch $b$ from buffer $\mathcal{B}$
\State $d = \frac{1}{|b|}\sum_{z_t \in b} \norm{f_\theta(s) - z_t}_2$ \label{alg:alg3:distance}
\If{$d < d_{\max}$} \label{alg:alg3:safety_check}
\State With a probability $\rho$:\textbf{ } \label{alg:alg3:prob}
\While{${Q^*_P}(s,a)<\underset{a'\in\mathcal{A}}{mean}{Q^*_P}(s,a')$}
\State Pick random action from $\mathcal{A}:$ $a_{0}=random(\mathcal{A})$ 
\EndWhile \label{alg:alg3:endwhile}
\EndIf
\State $a=a_{0}$
\end{algorithmic}
\label{alg:algorithm3}
\end{algorithm}
\section{Experiments and Results}
\subsection{Experiment Settings}
\begin{figure*}[h]
	\centering
	\begin{subfigure}[b]{0.3\textwidth}
		\centering
		\includegraphics[width=\textwidth]{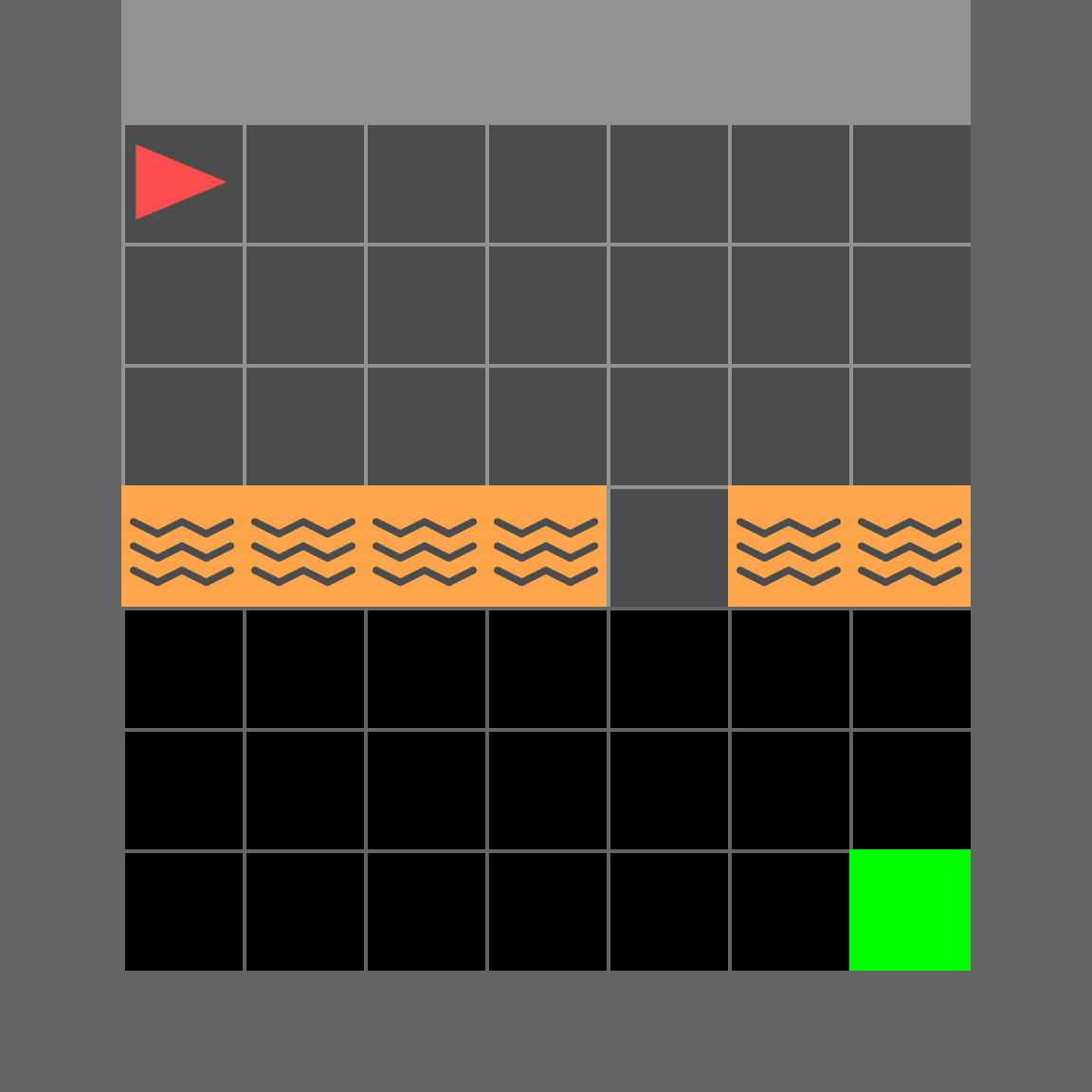}
		\caption[]%
		{{\small LavaCrossingS9N1}}    
		\label{fig:LavaCrossingS9N1}
	\end{subfigure}
	\hfill
	\begin{subfigure}[b]{0.3\textwidth}  
		\centering 
		\includegraphics[width=\textwidth]{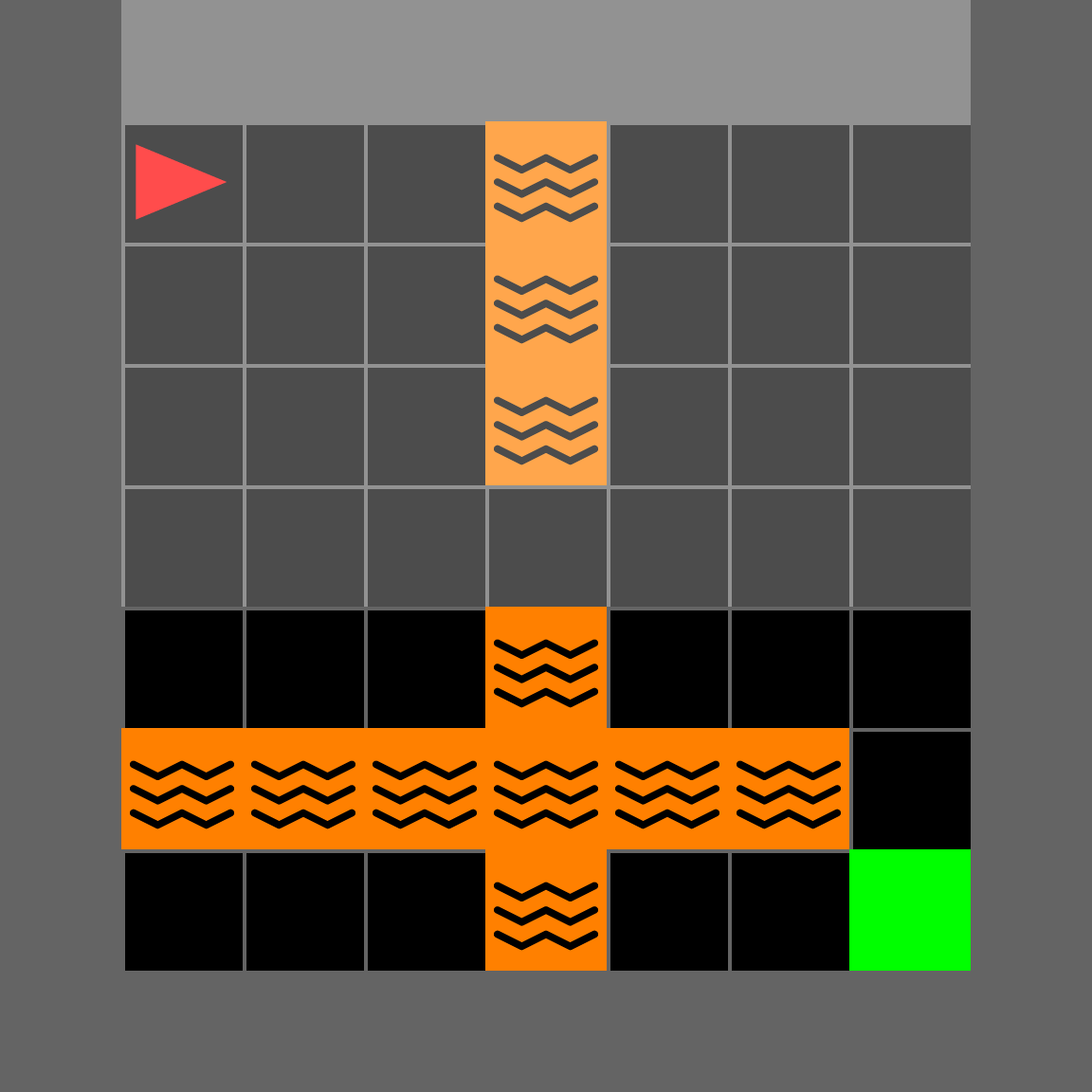}
		\caption[]%
		{{\small LavaCrossingS9N2}}    
		\label{fig:LavaCrossingS9N2}
	\end{subfigure}
	\hfill
	\begin{subfigure}[b]{0.3\textwidth}   
		\centering 
		\includegraphics[width=\textwidth]{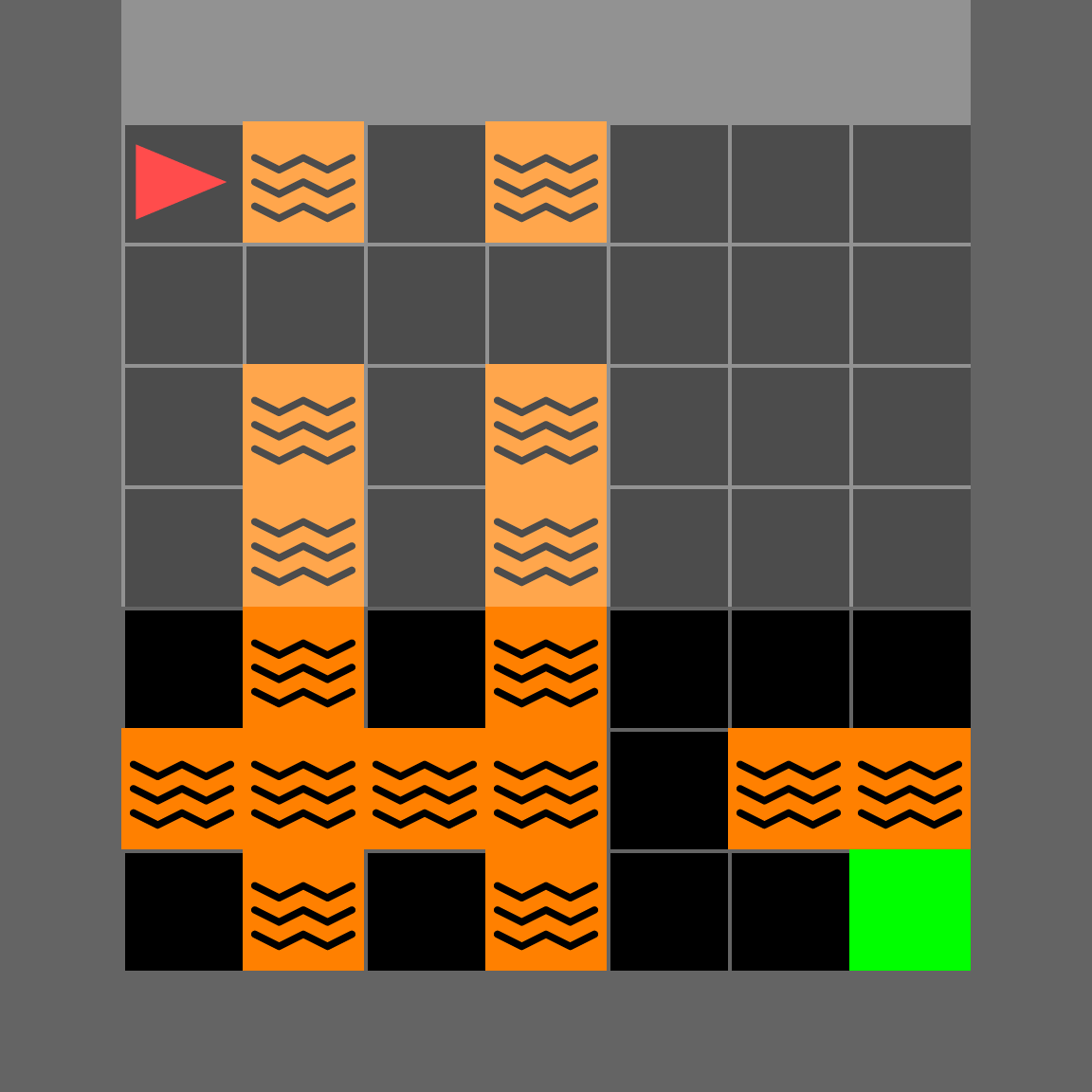}
		\caption[]%
		{{\small LavaCrossingS9N3}}    
		\label{fig:LavaCrossingS9N3}
	\end{subfigure}
	\caption[]
	{\small Minigrid Environments} 
	\label{fig:minigrid}
\end{figure*}

We evaluate our method on three safety environments from MiniGrid \cite{MinigridMiniworld23}, which are \textit{LavaCrossingSXNY} environments, where X and Y denote the size of the map and the number of crossings across lava to reach the goal respectively.
We use the full observation setting with RGB images input, i.e. the agent has a full view of the grid (Figure \ref{fig:minigrid}).
In these environments, the agent can only take three actions: turn left or right, which changes the direction the agent is facing, and move forward, which moves the agent one cell in the current direction. The agent starts from the red cell in the upper left corner and has to reach the green cell in the lower right corner, while avoiding stepping into lava cells, which leads to termination.
The stepping actions into the lava cells are considered unsafe actions.
The reward function is zero for all transitions except for the one that leads to the goal state, which follows a sparse-reward setting.
The agent has to tackle the problem of sparse rewards and safety together.

\textbf{Baselines.}
We present the results of our method in comparison with other methods as well as an analysis of the impact of state safety check on the performance of the agent. The baselines include the following methods and algorithms:
\begin{itemize}
    \item DQN: a Deep Q-Network agent \cite{mnih2015human} with prioritized experience replay \cite{schaul2016prioritizedexperiencereplay} and double Q-learning \cite{vanhasselt2015deepreinforcementlearningdouble}. The DQN agent also serves as the base agent for other baselines.
    \item DQN + AE: a Deep Q-Network agent that takes as input a latent vector of the observation encoded by an auto-encoder (AE).
    \item DQN + VAE: the same as DQN + AE but a Variational Auto-encoder is used (VAE) instead of an AE.
    \item DQN + Domain Priors: a DQN agent that uses transferable domain priors \cite{karimpanal2020} approach for safe exploration. The prior Q-network is trained by aggregating previously learned Q-learned functions in four environments: SimpleCrossingS9N1, SimpleCrossingS9N2, LavaCrossingS9N1 and LavaCrossingS9N2. The entropy threshold $t$ is set to $0.1$.
\end{itemize}
Our method, named DQN + Domain Priors + State Safety Check, further adds bias exploration behaviors upon the baseline DQN with Domain Priors.
All experiments are conducted with the same network architecture and hyper-parameters, if they are related.

\textbf{Model and hyperparameters.}
The auto-encoder trained with contrastive objective (equation \ref{eq:AEobj}) is composed of 3 hidden layers and 32 convolution filters of size $3 \times 3$ for each layer. The dimensionality of the latent space is 50. $\omega_1$ and $\omega_2$ is set to 100 and 0.01 respectively. To separate two groups of safe and unsafe states, we choose the margin $\alpha = 10$. As for training the policy, the input RGB image is normalized and then passed through a sequence of 3 convolutional layers with 32 filters each and decreasing strides of 4, 2 and 1 respectively. A rectified linear unit (ReLU) layer is applied after each convolutional layer. The output of the convolutional block is then flattened and fed into a full-connected neural network with a hidden layer of 512 units. Regarding the safety exploration module, the size of the buffer $\mathcal{B}$ and batch $b$ is 100 and 10 respectively, while the max distance $d_{\max}$ is 2.5. The probability of interfering with unsafe action $\rho$ is 0.95. All models are optimized using Adam with a learning rate of $2.5 \times 10^{-4}$.
Each method is trained with 200K, and 3 seeds.
The plot the mean reward and number of violations are presented in Fig. \ref{fig:episodic_returns} and Fig. \ref{fig:cum_num_violations}, respectively.

\subsection{Results}
\begin{figure}
    \centering
    \includegraphics[width=\linewidth]{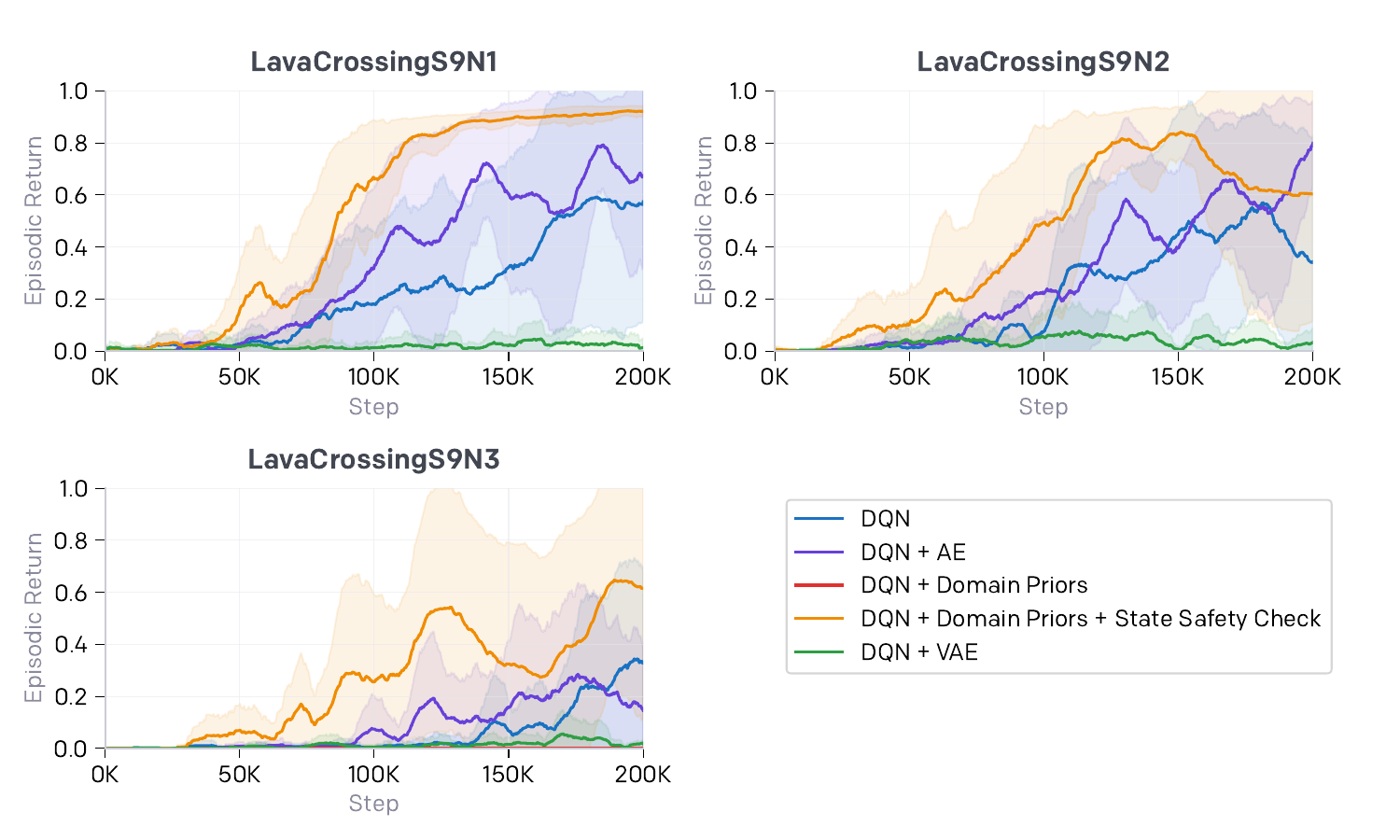}
    \caption{The episodic returns for 5 agent in on 3 MiniGrid environments, averaged over 3 seeds.}
    \label{fig:episodic_returns}
\end{figure}

\begin{figure}
    \centering
    \includegraphics[width=\linewidth]{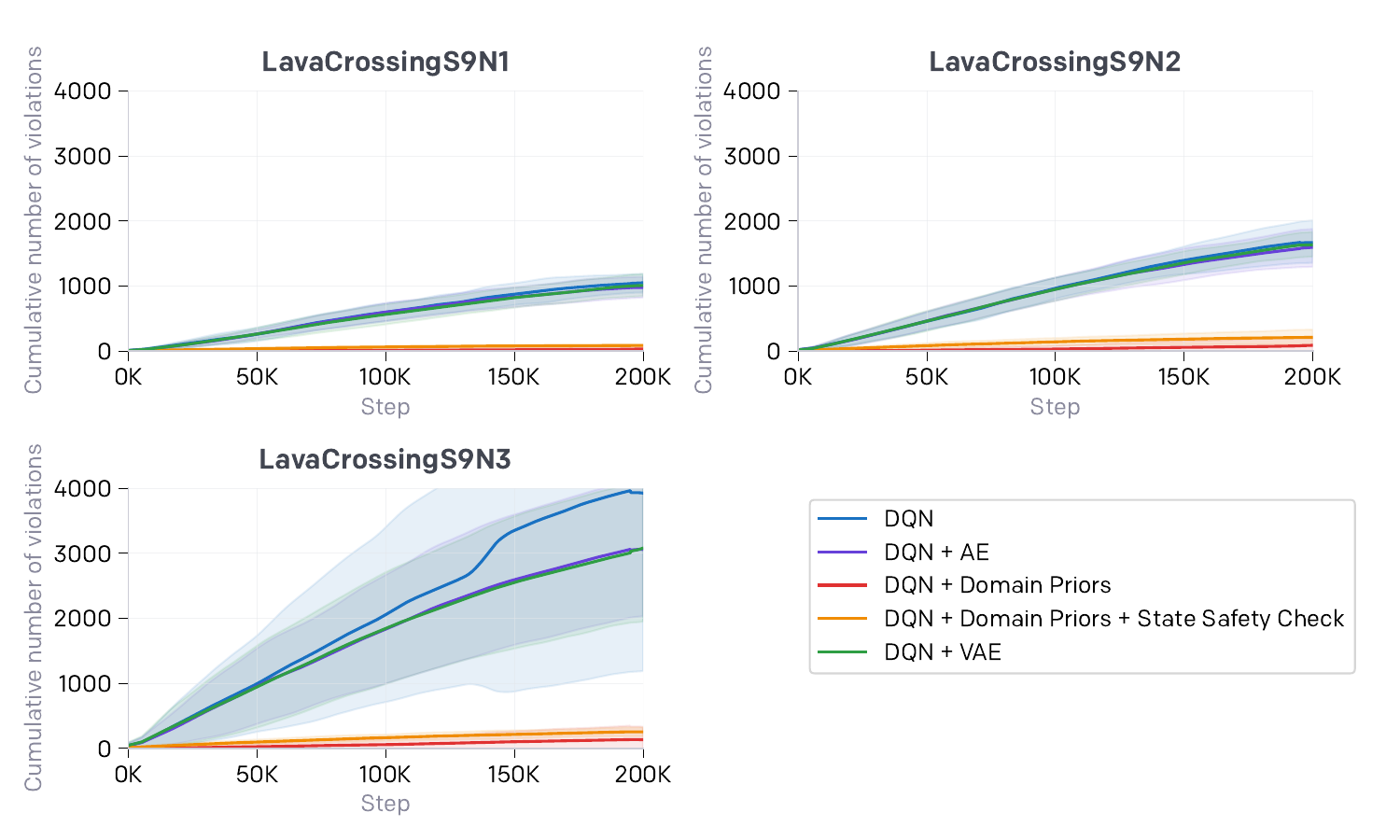}
    \caption{Performance of 5 agents in terms of the number of violations.}
    \label{fig:cum_num_violations}
\end{figure}
As for the cumulative episodic returns in Fig. \ref{fig:episodic_returns}, out method outperforms the rest in LavaCrossingS9N1 and LavaCrossingS9N3, albeit with high variance in S9N2 and S9N3. Noticeably, the original method with transferable domain priors shows the inability to complete the tasks, leading to zero reward. Our method is shown to effectively addresses this by selectively biasing exploration. The other methods are also able to solve the task; however, the rewards decrease in line with the difficulty of the environments. This can be explained by the number of violations, shown in Fig. \ref{fig:cum_num_violations}. Methods without safety mechanisms visit unsafe states at a much higher rate than our method and transferable domain prior method. As stepping into unsafe states terminates the episode, the former also has difficulty completing the tasks, especially in the ones where unsafe states are pervasive such as LavaCrossingS9N3. In these environments, the ability to avoid unsafe states are crucial in reach the goal. For a comparison of the latter i.e. methods with safety awareness, our method performs worse than the original method. This can be explained by the reward and safety tradeoff [citation]. As our method explores more to address the sparsity of the environment, the likelihood of visiting unsafe states also increases. In contrast, the original method prioritizes safety, which confines the agent to only a safe region, as we discuss in the following section.

\begin{figure}[h]
	\centering
	\includegraphics[width=\linewidth]{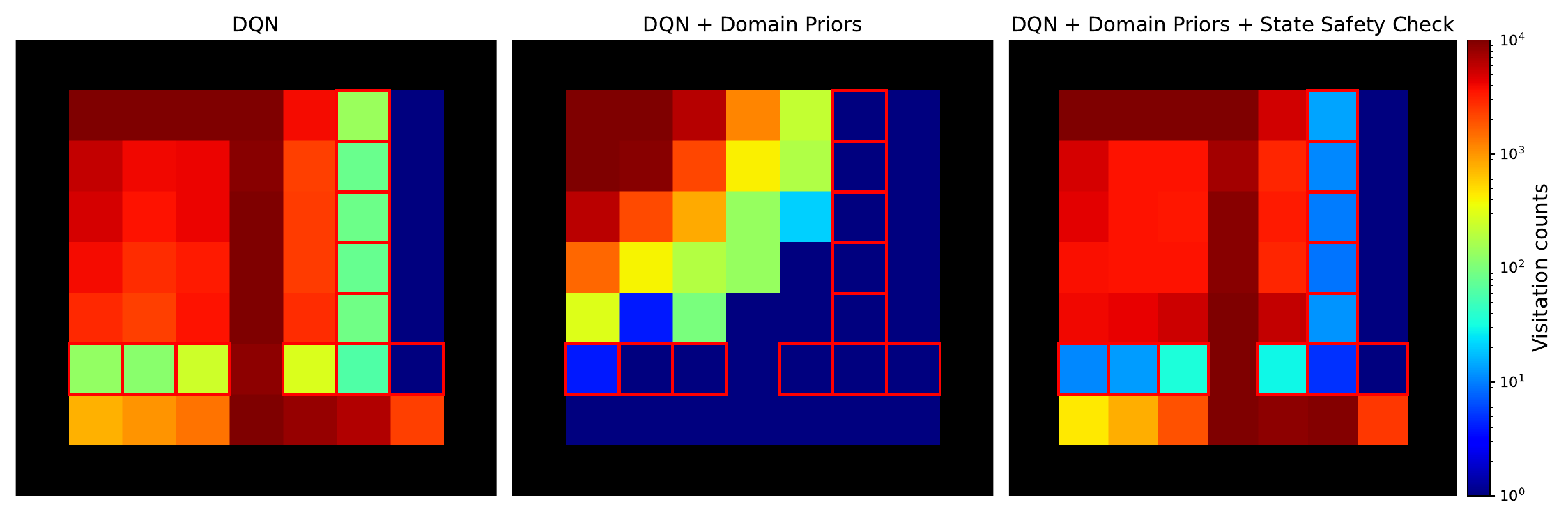}
	\caption{Exploration heatmap for DQN agent (left), transferable domain prior agent (middle), and our agent (right). The warmness of the color indicates how much the agent visits unsafe states.}
	\label{fig:exploration}
\end{figure}

\textbf{Analysis of the effects on exploration}. Figure \ref{fig:exploration} compares the extent of exploration in S9N2 environment for the vanilla DQN agent (left), the agent with transferable domain priors (middle) and our agent (right). These figures show a heatmap of the number of visitations per cell for each agent. The agent starts at the upper left corner cell and has to reach the lower right corner cell. The lava cells are marked with red border. The exploration behaviour of three agents are different. While the DQN and our agent are able to complete the task, the agent trained with transferable domain priors method fails, which explains the zero reward in Fig. \ref{fig:episodic_returns}. The reason is that exploration is biased in almost every states when the probability of forcing an agent to execute an alternative action is high. As a result, the agent keeps staying in the safe region around the starting point without risk exploring further. This can be seen in Fig. \ref{fig:exploration} (left), in which cells closer to the starting cell are visited more often. In contrast, the DQN agent and our agent exhibit similar exploratory behaviour in that all states are visited. However, our agent is much less likely to visit unsafe states, which is the desired behaviour.

\begin{figure}[h]
	\centering
	\includegraphics[width=\linewidth]{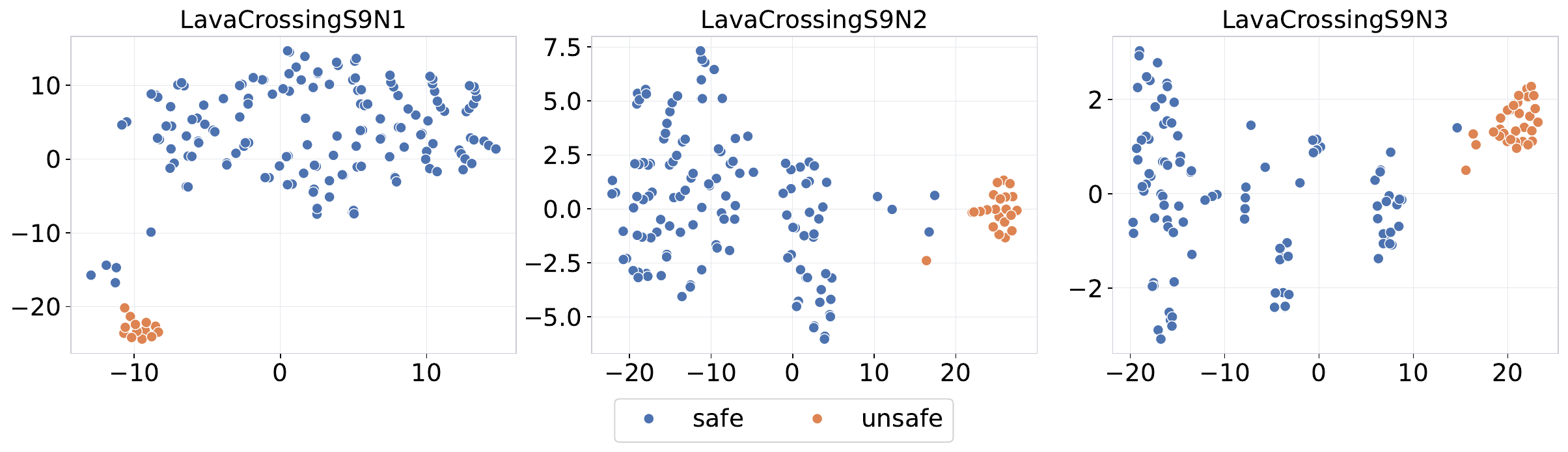}
	\caption{The learned embeddings of all observations in three environments.}
	\label{fig:tSNE}
\end{figure}

\textbf{Analysis of the learned latent space}. Figure \ref{fig:tSNE} illustrates the learned embeddings of observations corresponding to all states in the three above environments. The embedding network is only trained on LavaCrossingS9N1. Therefore, in this environment, the embeddings corresponding to safe and unsafe states are clearly separated. In other environments, as the observation space is similar, the embedding network is able to generalize well enough so that the agent can distinguish between safe and unsafe observations.

\section{Conclusion}
In this work, we propose a new method that builds on transferable domain priors and contrastive learning to maintain both safety and exploration in observation-based environments. Contrastive learning is employed to map observations to a latent space where the agent can distinguish safe and unsafe states, while transferable domain prior is used to choose safe actions. By decoupling state safety from action safety, exploration is only biased in safe states. Our experiments on MiniGrid environments show that our method is able to avoid safety violations while maintaining extensive exploration, even with sparse-reward signals.
From the results of this paper, it is possible to extend our method by varying the unsupervised latent learning loss other than contrastive loss. Moreover, as the safety latent distance threshold is fixed in our proposed method, it can be extended to an adaptive learning mechanism to improve stability.
%
%

\section*{Acknowledgment}
This work has been supported by VNU University of Engineering and Technology under project number CN24.13.

\bibliographystyle{unsrt}
\bibliography{main}

\end{document}